%% file: main.tex
\documentclass[11pt]{article}

\usepackage[T1]{fontenc}
\usepackage[english]{babel}
\usepackage[scaled=.96,osf]{XCharter}
\usepackage[xcharter]{newtxmath}
\usepackage{multirow}

\usepackage{amsmath,amssymb,amsthm,mathtools}
\usepackage{algorithm}
\usepackage{algorithmic}

\usepackage[margin=1in]{geometry}
\usepackage[dvipsnames]{xcolor}
\usepackage{graphicx,tikz,cancel,comment,mdframed}
\usepackage[inline]{enumitem}
\usepackage{hyperref}   
\usepackage{booktabs}
\input{macros}

\usepackage[round]{natbib} 
\title{Interpretable \& Fine-grained Novelty Detection via LLM Embeddings}
\author{Nico Arsenio LIMAN \\
        National University of Singapore \\
        \texttt{e1374368@u.nus.edu}
        \and
        Limsoon WONG\\
        National University of Singapore \\
        \texttt{limsoon@nus.edu.sg}}
        
\begin{document}
\maketitle
\section{Abstract}
Novelty in research has often been considered an integral part of good research. As the number of papers rises, it becomes increasingly challenging to find novel works relevant to ones work. Current research focuses on LLM judgement, which functions as a black box that is difficult to tune and interpret. An alternative method is that which measures the semantic similarity of research works with respect to its peers. Nevertheless, most works of this nature have adopted a relatively coarse approach. We provide a more tractable definition of novelty and provide a method for detecting novelty without time lag that we argue to computationally mimic our perception of novelty.
\section{Introduction}\seclabel{intro}
\subsection{Significance of Novel Papers and its definition} 
Revolutionary papers have often been lauded as novel in retrospect. The polymerase chain reaction (PCR), for example, is now the standard method for amplifying a specific DNA sequence \citep{saiki1985enzymatic, mullis1986specific, saiki1988primer}, yet it departed from the dominant approach of its time. Until 1985, the method of choice for obtaining copies of a specific sequence was cloning, in which the sequence is inserted into a plasmid and grown in bacteria \citep{doi:10.1073/pnas.78.11.6613, epo2003pcr}. Interestingly, PCR was not without precedent. \citet{kleppe1971repair} had already described primer-directed replication of a short DNA segment and outlined repeating it, but did not test the repeated process. The same group later abandoned this line of work \citep{panet1974linkage, epo2003pcr}, and the components PCR relies on were available long before it was developed \citep{mullis1990unusual, epo2003pcr}. 
There is also some quantitative evidence for novelty being a key ingredient for revolutionary papers. In the work of \cite{doi:10.1126/science.1240474}, they have found that top 5\% of works by citation in a given field tend to have several atypical references with respect to its peers which suggests that impactful works tend to borrow slightly from different fields.

In this work, we define two types of novelty and propose a method for detecting novel works. A work exhibits \textit{problem novelty} when the problem it addresses differs substantially from neighbouring problems in the literature. When the problem is relatively common, a work may instead exhibit \textit{approach novelty}, where its method of solving the problem departs substantially from established approaches. Importantly, we treat similarity as graded rather than binary. Consider a paper that presents the first method for finding the roots of a quadratic equation, and a later paper that presents the first method for a cubic. From a problem perspective, the second paper is not highly novel as it is a natural extension from the first. From an approach perspective however, the method for quadratics does not translate nicely to cubics and require new insight. A single novelty score would conflate these two judgements, whereas separating problems from approaches distinguishes them.

In short our contributions is as such:  
\begin{itemize}
    \item We propose a tractable definition of novelty for research papers
    \item We present an interpretable algorithm for quantifying/detecting novel works 
\end{itemize}

\subsection{Methods for Novelty Detection}

\subsubsection{LLM as a Judge}
Recent studies focuses on using large-language models (LLMs) directly as judges of novelty, either by prompting them alone or by grounding their judgements in retrieved literature. \citet{si2024llms} converted ICLR submissions into standardised project proposals and ranked them with a Swiss-system tournament of pairwise LLM comparisons, with Claude-3.5-Sonnet reaching 71.4\% accuracy in predicting acceptance. \citet{lu2024aiscientist} built a novelty check into the AI Scientist, in which an LLM searches Semantic Scholar for relevant works to judge novelty. Several works add retrieval to the judging step. \citet{lin-etal-2025-evaluating} proposed RAG-Novelty, which supplies the average publication date of retrieved papers as context and outperformed prompting strategies such as chain-of-thought, self-consistency and multi-agent discussion on their SchNovel benchmark. \citet{shahid-etal-2025-literature} proposed the Idea Novelty Checker, which combines keyword and snippet retrieval, embedding-based filtering and facet-based LLM re-ranking, and reported about 13\% higher agreement with expert labels than earlier approaches. OpenNovelty \citep{zhang2026opennovelty} takes an agentic approach, grounding each novelty claim in retrieved papers with citations and evidence snippets. \citet{afzal-etal-2026-beyond} evaluated LLM novelty assessments on 182 ICLR 2025 submissions, reporting 86.5\% alignment with reviewers' reasoning but only 75.3\% agreement on the final novelty conclusion.

Other work questions how reliable these judges are. NovBench \citep{wu2026novbench} scores LLM-written novelty reviews against 1,684 paper--review pairs and finds that current models have a limited understanding of scientific novelty. RINoBench \citep{schopf-farber-2026-idea} finds that LLM reasoning closely resembles human rationales, yet the final novelty judgements diverge from expert judgements, and \citep{schopf-farber-2026-idea}  trace this to a systematic bias towards medium novelty ratings. \citet{sternlicht2026oldideas} show that small changes to the evaluation setup can flip a judge's verdict on more than half of identical idea pairs, and note that new ideation systems typically ship their own ad hoc LLM judge. RQ-Bench \citep{sinhahajari2026limits} finds that LLM judges rate model-generated research questions as highly novel even when domain experts prefer the author-written ones. \citet{wang2025enablingaiscientistsrecognize} likewise found that an LLM judge without retrieval (Sonnet-3.7) performed close to chance (AUROC $\approx 0.5$), and that even with retrieval, its performance dropped sharply from computer science to biomedicine (AUROC 0.81 to 0.62).

These findings point to a central weakness of LLM judges: they are unreliable, and we cannot determine why. A judge can produce a rationale that reads like an expert's while its final score disagrees with experts \citep{schopf-farber-2026-idea}, so the stated reasoning does not reveal how the verdict was reached. The same judge can also reverse its verdict under small changes to the evaluation setup \citep{sternlicht2026oldideas}, without any indication of what caused the change. Without knowing what the model treats as novel, we can neither diagnose its errors nor decide how to correct them. This matters all the more because novelty is subjective and depends on who is judging; even expert annotators can disagree \citep{schopf-farber-2026-idea,sinhahajari2026limits}.
\subsubsection{Embedding-Based Detection}
A separate line of work measures novelty without LLM judgement, using distances in an embedding space. Early approaches focus on recombination. \citet{shibayama2021measuring} consider a paper novel if it cites semantically distant references: they embed each reference title with scispaCy word embeddings and summarise the pairwise distances between references. \citet{yin2023identify} instead measure element novelty, the distance of a document from earlier literature, using a word embedding model trained on titles and abstracts across several scientific fields. Both measures were validated against novelty scores self-reported by authors, and \citet{shibayama2021measuring} also showed that their measure predicts future citations. \citet{su2024manyheads} use Historical Dissimilarity as the mean Euclidean distance between an abstract and its five most similar abstracts in a historical corpus, and combine it with a contemporary counterpart and a citation-based impact score into an Overall Novelty metric. Later work replaced static word vectors with transformer encoders such as BERT, which take word order and context into account.\citet{culbert2025originality} adapt Divergent Semantic Integration, an originality measure from creativity science, to 51,200 scientific titles and abstracts, using the distances between contextualised embeddings of sentences within each text. \citet{wang2025enablingaiscientistsrecognize} argue that absolute distances are not comparable across fields with different publication densities. Their Relative Neighbor Density (RND) instead scores an idea by the percentile of its local density among the densities of its nearest neighbours. RND matched an absolute-distance baseline within single domains but kept an AUROC of 0.795 on a mixed computer science and biomedical test set, where the baseline fell to 0.362. Finally, \citet{wu2025automated} take a supervised approach, fine-tuning pre-trained language models such as BERT to predict method novelty from LLM-generated summaries of a paper's methods and novelty-related sentences from its peer reviews.
 
The work closest to ours is \citet{wu2025economics}, which decomposes each paper into facets before embedding. An LLM first condenses each full-text economics paper into a structured summary, then extracts five facets from it: an overall summary, the topic, the research question and conclusions, the methodology, and the theory. Each facet is embedded separately, and the paper's isolation from earlier work on each facet is measured with nearest-neighbour distances and local density. These measures are pooled into two scores: spatial novelty, a paper's isolation from prior work, and temporal novelty, how quickly its neighbourhood filled with new work before publication. The former predicts disruptive impact and the latter citation counts.
 
Nevertheless, we argue that several choices in this design introduce confounding factors. The topic facet asks the LLM to name a single subfield, forcing each paper into a discrete field even though research often spans fields; a paper between two fields must thus be filed under one of them. The research question facet includes the paper's findings, which reflect how an experiment turned out rather than whether the idea was new, since the same approach can yield strong or weak results. The prompts also ask the LLM to describe what makes each paper's findings, methods and theory new, and the summary from which the facets are extracted includes a disruptiveness rating, so the embedded text partly carries the LLM's own novelty judgement. Finally, each composite score is the first principal component of metrics pooled across all five facets, so the scores do not show which aspect of a paper makes it novel, and the scores are validated against citations and disruption rather than against novelty labels.

\section{Proposed Method}\seclabel{method}

We first describe the general method and the purpose of each step, then the particular instantiation we use to validate it. Following \citet{mysore2021csfcube}, we use the term \emph{facets} to refer to rhetorical elements of a paper. They defined three facets (background, method and result), whereas we use two: the \emph{problem}, meaning what the paper sets out to solve, and the \emph{approach}, meaning how it solves it. For each paper $i$, we also define two sets of papers. The \emph{reference set} $\mathcal{R}_i$ contains the papers published before $i$, against which its problem and approach are compared. The \emph{cohort} $\mathcal{C}_i$ contains the papers published in the same period as $i$, which helps establish the baseline.

Firstly, an LLM reads the paper and writes a short description of each facet: one sentence stating the
problem and one stating the approach.  They should describe the paper accurately while removing confounding factors, such as background information that many papers share, and noise from the wording of the paper itself: its length, and descriptive language such as ``novel'' or ``interesting'' that reflects the authors' opinion of their work rather than what the work does.

Secondly, these descriptions are embedded in a latent space where similar descriptions lie close together under a chosen distance metric. For example, a paper on solving quadratic equations should lie closer to one on solving cubic equations than to one on protein folding. How this can be achieved will be apparent when discussing our particular instantiation. 

Lastly, we find the $k$ papers in $\mathcal{R}_i$ whose problems are closest to the problem of paper $i$, and compare paper $i$ with these $k$ neighbours. For each comparison, each cohort paper’s distance is measured to its own $k$ neighbours in its own reference set. Paper $i$ exhibits \textit{problem novelty} when its mean problem distance to its $k$ neighbours is larger than that of a given proportion of the papers in its cohort $\mathcal{C}_i$. When its mean problem distance is lower, a similar problem has already been addressed, and paper $i$ exhibits \textit{approach novelty} if its mean approach distance to the same $k$ neighbours is larger than that of a given proportion of the papers in its cohort.

\section{Evaluation}
We evaluate our method in several ways. We first describe the instantiation of the method used in all of the evaluations, and then present each evaluation in turn.

\subsection{Instantiation}\label{method:inst}
\paragraph{Facet extraction.} For each paper in a cohort and each earlier paper in its reference set, we give the full text to DeepSeek-V4.1-Flash-1017 and obtain one sentence describing the problem and one describing the approach. We restrict each description to one sentence because the embedding model is trained to provide embeddings for one sentence. The two descriptions are produced by separate prompts. The problem prompt asks for the research problem in one sentence and instructs the model to leave out how the problem is addressed. The approach prompt asks for one sentence that describes only the research approach. Neither prompt asks the model whether the work is new or important, so the authors' claims of novelty are less likely to be carried into the descriptions. Both prompts are given in Appendix~\ref{app:prompts}. We use the model as released (fp8 weights, with fp4 experts), served with vLLM 0.32.0 \citep{kwon2023vllm} and an fp8 KV cache, with thinking mode at the default reasoning effort and a temperature of 0. The model's context window of one million tokens fits every paper in full, so no text is truncated.

\paragraph{Embedding.}  Each problem description and each approach description is embedded separately with \texttt{aspire-contextualsentence-singlem-biomed}, the single-match variant of Aspire \citep{mysore2022aspire} trained on biomedical papers. A description of $n$ tokens is passed through the encoder, which returns one vector per token, $\mathbf{X} \in \mathbb{R}^{n \times h}$, where $h = 768$ is the hidden size of the encoder. As descriptions differ in length, $n$ varies between descriptions, we mean-pool over the tokens to obtain a single vector, \textbf{X} of fixed size.
\begin{equation}
    \mathbf{X} = \frac{1}{n} \sum_{t=1}^{n} \mathbf{x}_{t} \in \mathbb{R}^{h},
\end{equation}
where $\mathbf{x}_{t}$ is the embedding of the $t$-th token. This is the same pooling Aspire uses during training, where each sentence of an abstract is represented by the mean of its token embeddings. The problem and approach embeddings of paper $i$ are denoted $\mathbf{p}_i$ and $\mathbf{a}_i$.

Aspire suits our purpose because of how it is trained. It is trained from co-citations: papers cited together in the same sentence of another paper. More importantly, rather than pulling the whole abstracts of co-cited papers together, it pulls together only the pair of sentences, one from each abstract, that best matches the citing sentence. Since citing authors often cite papers together because they share a specific element, such as the same problem or the same method, the model learns to place sentences describing that shared element close together, without forcing the rest of the two papers to match. Two papers that address the same problem with different methods therefore have similar problem embeddings but not necessarily similar approach embeddings, which is the separation our scores rely on.

 \paragraph{Novelty scores.} Given the problem and approach embeddings of each paper, we measure distances with the L2 distance $d(\mathbf{u}, \mathbf{v}) = \lVert \mathbf{u} - \mathbf{v} \rVert_2$.For each paper $i$ in a cohort, the reference set $\mathcal{R}_i$ contains all papers published before $i$ in the window defined above, which number in the thousands to tens of thousands. We exclude from $\mathcal{R}_i$ any paper with the same corresponding author as $i$, which we take to indicate the same research group. A group's earlier papers on the same problem are usually its own preliminary or preceding work, and comparing against them would measure how far the group has moved from its own past rather than how far the paper departs from the rest of the field. From the remaining papers, we select the $k = 20$ whose problems are closest to the problem of $i$; we denote this set $\mathcal{K}_i$. All scores are computed over $\mathcal{K}_i$ only. The problem score is the geometric mean distance from $i$'s problem to the problems of these 20 papers,
\begin{equation}
    N_P(i) = \left( \prod_{j \in \mathcal{K}_i} d(\mathbf{p}_i, \mathbf{p}_j) \right)^{1/k},
\end{equation}
and the approach score is the geometric mean distance from $i$'s approach to the approaches of the same $k$ papers,
\begin{equation}
    N_A(i) = \left( \prod_{j \in \mathcal{K}_i} d(\mathbf{a}_i, \mathbf{a}_j) \right)^{1/k}.
\end{equation}

\paragraph{Choice of $k$.} We use a relatively large $k$ because the score is intended as a proxy for sparsity: how few earlier papers lie near a paper's problem or approach. Distance to the closest neighbour does not always capture this. For example, two papers may lie at the same distance from their closest precedents, one at the centre of a crowded area and the other in a region where only a few earlier papers exist. A small $k$ scores them the same, whereas a larger $k$ distinguishes them, since the remaining neighbours of the first are close and those of the second are far. Only the second explores a direction that few others have taken, as with PCR, whose approach had been described earlier but not pursued \citep{kleppe1971repair, epo2003pcr}. A large enough $k$ also accommodates simultaneous discovery. Important discoveries are often made independently by more than one group around the same time \citep{merton1961singletons, hill2025scooped}, and since the reference set includes earlier papers from the same year, each such paper may have the others among its neighbours. With a small $k$, each would be marked conventional because of the others; with a larger $k$, a few concurrent papers lower the score only slightly.

\subsection{Study on Howard Hughes Medical Institute Investigators}\label{eval:HHMI}
In our first experiment, we examine papers by investigators of the Howard Hughes Medical Institute (HHMI). \citet{HOLMSTROM1989305} describes innovative projects as risky, unpredictable, long-term, labour-intensive and idiosyncratic. Funding schemes that demand short-term results are therefore poorly suited to them. Grants from the National Institutes of Health (NIH), for example, are subject to short review cycles, pre-defined deliverables and renewal policies that are unforgiving of failure. HHMI, by contrast, tolerates early failure, rewards long-term success and gives its investigators considerable freedom to experiment \citep{NBERw15466}, the combination that theory predicts best motivates innovation \citep{https://doi.org/10.1111/j.1540-6261.2011.01688.x}. Studying investigators appointed between 1993 and 1995, \citet{NBERw15466} found that they produced high-impact papers at a much higher rate than similarly accomplished NIH-funded scientists, and that the direction of their research changed in ways that suggest the program led them to explore novel lines of inquiry. HHMI papers therefore provide an external group that we expect to be more novel than comparable work.

Numerically, we can also see that 36 scientists supported by HHMI have won the Nobel prize.

\paragraph{Data.} We use the open-access subset of PubMed Central (PMC), which provides full text in bulk. It contains
50{,}536 articles published between 1970 and 1995 (31{,}926 from 1970--1989 and 18{,}610 from 1990--1995). We keep
research articles and brief reports with at least 2{,}000 characters of full text, and drop 12 papers for which
extraction failed, leaving 38{,}906 papers. For each paper, we extract the problem and approach sentences from the full
text and embed them as described in Section~\ref{sec:method}.

The 13{,}350 papers published between 1990 and 1995 are the target papers; $\mathcal{D}_Y$ denotes the target papers published in year $Y$. Each target paper is compared with every paper in the corpus published before it, back to 1970: a paper from 1993, for example, is compared with all papers from 1970 to 1992 and with those from 1993 that appeared earlier. A paper from the same year counts as earlier only when the publication months (and, within the same month, the days) of both papers are known from the PMC citation record and the candidate's is earlier. We also exclude the authors' own earlier work: a candidate is dropped if it shares a last or corresponding author with the target paper (matched on surname and first initial). $N_P(i)$ is the geometric mean problem distance from paper $i$ to its $k = 20$ nearest remaining candidates by problem distance, and $N_A(i)$ is the geometric mean approach distance to the same 20 papers.

\paragraph{HHMI papers.} We use each year $Y \in \{1990, \dots, 1995\}$ as a target year. We compiled the names and
appointment years of all HHMI Investigators from archived pages of the HHMI website (949 Investigators, 326 of whom held
the appointment in at least one year between 1990 and 1995); other HHMI programmes are not included. A paper in
$\mathcal{D}_Y$ is labelled an HHMI paper if:
\begin{itemize}
    \item its last author or a corresponding author matches an Investigator on this list (surname and first initial)
    \item $Y$ falls within that Investigator's appointment , and
    \item the paper mentions the Howard Hughes Medical Institute in an affiliation, footnote, acknowledgement or the text.
\end{itemize}

This gives 228 HHMI papers. We set aside the 528 papers that mention HHMI or list a current Investigator as a co-author without
meeting all three conditions (``other HHMI-linked''). The remaining 12{,}594 papers form the comparison group.

\paragraph{Within-year thresholds.} While each paper's neighbours are drawn from all earlier papers, its scores are ranked only against those of papers published in the same year. For a target year $Y$ and a percentage $q$, let $\tau_P(Y,q)$ and $\tau_A(Y,q)$ be the $(100-q)$th percentiles of $N_P$ and $N_A$
over $\mathcal{D}_Y$ (all target papers of that year, in every group). A paper $i \in \mathcal{D}_Y$ exhibits problem
novelty if $N_P(i) > \tau_P(Y,q)$, that is, if its problem distance is in the top $q\%$ of its year. It exhibits
approach novelty if $N_P(i) < \tau_P(Y,q)$ and $N_A(i) > \tau_A(Y,q)$: its problem is no more distant than usual, but
its approach is in the top $q\%$ of its year. A paper is novel if it exhibits either type. We report results for $q \in \{50, 20\}$.

\paragraph{Test.} Because thresholds are set within each year, $q\%$ of each year's papers exhibit problem novelty by
construction. The share exhibiting approach novelty is not fixed by construction; over all target papers it is 14.1\%
at $q = 50$ and 9.9\% at $q = 20$. If our measure captures novelty, and HHMI investigators produce more novel work
\citep{NBERw15466}, then HHMI papers should be flagged at a higher rate than comparison papers. For each type of
novelty, we compare the share of HHMI papers flagged with the share of comparison papers flagged, pooled over
1990--1995, using a two-sided Fisher's exact test.

\paragraph{Results.} Table~\ref{tab:hhmi} reports the proportion of HHMI, other HHMI-linked and comparison papers
flagged as novel. At $q = 50$, 50.9\% of HHMI papers exhibit problem novelty, compared with 50.0\% of comparison papers
($p = 0.84$). For approach novelty, the rates are 23.2\% and 13.9\% ($p = 1.6 \times 10^{-4}$), and 74.1\% and 63.9\%
of papers are novel by either criterion ($p = 0.0013$). At $q = 20$, the rates are 22.4\% and 20.0\% for
problem novelty ($p = 0.40$), 17.1\% and 9.7\% for approach novelty ($p = 6.7 \times 10^{-4}$), and 39.5\% and 29.8\%
for either ($p = 0.0021$). The difference in approach novelty is 9.3 percentage points at $q = 50$ and 7.4 at $q = 20$.
Other HHMI-linked papers do not differ from the comparison group at either threshold ($p \ge 0.12$).
\begin{table}[H]
\centering
\caption{Share of papers flagged as novel, 1990--1995. $p$: two-sided Fisher's exact test against the comparison
group.}
\label{tab:hhmi}
\begin{tabular}{rlrrrrr}
\toprule
 & & \multicolumn{2}{c}{HHMI ($n = 228$)} & \multicolumn{2}{c}{Other HHMI-linked ($n = 528$)} & Comparison \\
\cmidrule(lr){3-4}\cmidrule(lr){5-6}
$q$ & Novelty & \% & $p$ & \% & $p$ & ($n = 12{,}594$) \% \\
\midrule
50 & Problem  & 50.9 & 0.84                 & 49.2 & 0.76 & 50.0 \\
   & Approach & 23.2 & $1.6 \times 10^{-4}$ & 16.3 & 0.12 & 13.9 \\
   & \textbf{Either} & \textbf{74.1} & \textbf{0.0013} & 65.5 & 0.46 & \textbf{63.9} \\
\midrule
20 & Problem  & 22.4 & 0.40                 & 18.0 & 0.27 & 20.0 \\
   & Approach & 17.1 & $6.7 \times 10^{-4}$ & 11.6 & 0.18 &  9.7 \\
   & \textbf{Either} & \textbf{39.5} & \textbf{0.0021} & 29.5 & 0.92 & \textbf{29.8} \\
\bottomrule
\end{tabular}
\end{table}

\subsection{Case Studies and Failure Modes}\label{eval:case_studies}
We examine whether the model's judgements of novelty agree with ours on a small set of recent papers. We fixed the area to that of \citet{Mok2020}, which introduced CRISPR-free base editing of mitochondrial DNA.

\paragraph{Setup.} The setup follows the previously described instantiation (Section ~\ref{method:inst}). Nevertheless, the cohort and reference is built differently with respect to the previous experiment (Section ~\ref{eval:HHMI}). The cohort is drawn from the 7{,}133 papers published in 2020 that share at least one MeSH (Medical Subject Heading) major topic with \citet{Mok2020} and are journal articles with a PubMed abstract that are not reviews, comments, editorials or letters. We draw 300 of these at random and keep the first 100 with a usable full text in the PMC open-access subset (research articles or brief reports with at least 2{,}000 characters of text). We chose works published in 2020 mainly because it postdates the data used to train the embedding model (the S2ORC release of September 2019). Embedding all earlier literature was not feasible, so the pool of earlier papers combines three sources: papers from 2018--2019 sharing a MeSH major topic with \citet{Mok2020} (14{,}006 papers), the references of \citet{Mok2020} (44 of 64 kept), and the references of the cohort papers, all subject to the same filters and to having a usable full text, for 6{,}630 papers in total. Neighbours are selected such that they do not share the same corresponding author, similar to the previous experiment, with $q = 50$.

\paragraph{Blind test.} Each cohort paper falls into one of four groups according to whether it lies above or below the cohort median on each axis. From each group we took the three papers furthest into that corner, that is, with the largest $\min(|p_P - 50|, |p_A - 50|)$, where $p_P$ and $p_A$ are the paper's percentiles of $N_P$ and $N_A$. This gives 12 papers, six above and six below the median on each axis, lying at the 2nd–29th and 70th–98th of $N_P$; 2nd–36th and 76th–98th percentiles of $N_A$. The papers were presented in random order to the annotator (the first author), who knew only that each axis had six novel and six non-novel papers, and saw no groups, distances or percentiles. For each paper, the annotator read the full text and the problem and approach descriptions of the paper and its 10 nearest neighbours (all 20 were available separately), and labelled its problem and its approach as novel or not novel. Groups were revealed only after all 12 had been labelled.

\paragraph{Results.} The two problem-axis disagreements were papers the model placed above the median (at the 78th and 97th percentiles) that the annotator labelled not novel. On review, the annotator considered both borderline: their problems extend existing lines of work in ways that could reasonably be judged either way. The latter which was in the 97th percentile is evaluated a bit deeper in Appendix ~\ref{app:blind_ext} .The three approach-axis disagreements all come from the group with a high $N_P$ and a low $N_A$, where the annotator saw novelty that the model did not. When a paper’s problem is novel, its nearest neighbours address only partly similar problems, so their approaches are a weak reference for what a standard approach to this problem would be. 

\begin{table}[H]
\centering
\caption{Agreement between the annotator and the model by group (3 papers each).}
\label{tab:blind-mok-groups}
\begin{tabular}{lccc}
\toprule
Group & Problem & Approach & Both \\
\midrule
Low $N_P$, low $N_A$   & 3/3 & 3/3 & 3/3 \\
Low $N_P$, high $N_A$  & 3/3 & 3/3 & 3/3 \\
High $N_P$, high $N_A$ & 2/3 & 3/3 & 2/3 \\
High $N_P$, low $N_A$  & 2/3 & 0/3 & 0/3 \\
\midrule
All                    & 10/12 & 9/12 & 7/12 \\
\bottomrule
\end{tabular}
\end{table}

\section{Discussion \& Limitation}
\subsection{Discussion}
\paragraph{Relation to How People Judge Novelty \& Its Dimensions}

Our definition, novelty as distance from familiar work, is close to an intuitive notion of novelty: a paper strikes us as novel when its problem or approach is unlike what we already know. When people judge novelty this way, however, the judgement is subjective in three respects.

Firstly, any one person knows only a small fraction of the relevant literature, so a paper may seem new simply because the reader has not encountered similar work. Additionally, experts know different parts of the literature well depending on their area of expertise, so the same paper may seem novel to one reader and familiar to another. \citet{shahid-etal-2025-literature} observed this: experts judging the novelty of research ideas reached only moderate agreement (Cohen's $\kappa = 0.64$), in part because they drew on their own broader knowledge rather than the papers they were given.

Secondly, studies of how researchers judge originality distinguish several dimensions, such as new hypotheses or theories, methods, data and findings \citep{doi:10.1177/030631299029005004, guetzkow2004originality}, and disciplines weight these differently \citep{guetzkow2004originality}. A single judgement of novelty does not show which dimension a reader had in mind, so two readers may disagree because they weight the dimensions differently. \citet{shahid-etal-2025-literature} also note that their experts differed on how subtle contributions should count.

Thirdly, novel work often seems obvious once it is known. People who know how something turned out tend to judge that outcome as more predictable than it was \citep{fischhoff1975hindsight}, so a reader who is already familiar with a piece of work, or with the work that followed it, may underrate how new it was when it appeared.

Our measure keeps the intuition but attempts to address all three sources of subjectivity. Instead of any one reader's knowledge, every paper is compared with the same body of prior work, so its score does not depend on who is judging it. Instead of a single judgement, problem novelty and approach novelty are measured separately, so the score shows which aspect of a paper is new. Lastly, given that our model makes no direct judgement of novelty, hindsight bias would not occur. 

\paragraph{Novel Approach and Expertise}
Researchers know the problems and approaches near their own work best, and familiarity with an established solution can make alternatives harder to see. People tend to reuse a solution that has worked before even when a simpler one is available \citep{luchins1942mechanization}, and domain expertise can itself fix problem solvers on familiar approaches \citep{wiley1998expertise}. An effective approach may therefore exist in a distant field and go unnoticed by those working on a problem.
 
This would be captured by approach novelty. Because a paper's approach is compared only with the approaches of earlier papers that address similar problems ($\mathcal{K}_i$), an approach that is well established elsewhere but new to the problem is scored as novel. Such work may not introduce a new technique, but it is a novel application of a known one, and for those working on the problem the approach is new. Separating problems from approaches also allows these two cases to be distinguished: comparing a paper's approach with earlier approaches to the same problem identifies a novel application, while comparing it with earlier approaches to any problem would identify a novel technique.

Expertise need not lead to fixation. \citet{hatano1986two} distinguish routine experts, who apply known procedures efficiently, from adaptive experts, who understand a domain well enough to recognise where those procedures fall short and to adapt or replace them. The HHMI results may also be read in terms of how researchers engage with the models through which they work. A researcher may reason \emph{within} an accepted model of a problem, taking it as given, or reason \emph{about} the model itself, recognising its assumptions, judging whether it still fits, and adapting or replacing it. This distinction parallels that between single-loop learning, which corrects errors under existing assumptions, and double-loop learning, which questions the assumptions themselves \citep{argyris1978organizational}. HHMI papers are no more likely than others to address new problems, but are substantially more likely to bring a new approach to a familiar problem (Table~\ref{tab:hhmi}). This pattern is consistent with what \citet{hatano1986two} call adaptive expertise: experts who understand a domain well enough to recognise where established procedures fall short, and who adapt or replace them, in contrast to routine experts who apply known procedures efficiently but without change.

\subsection{Limitations}
\paragraph{Findings and Novelty}
Our framework does not measure novelty in a paper's findings as we expect this to be harder than measuring novelty in problems or approaches. This requires judging how unexpected a finding is given the problem and the approach used, and it is not yet clear how this should be defined. We also believe that novel findings are confounded with novel approaches: a new approach is more likely to produce results that differ from earlier work, so a measure of novel findings would partly measure approach novelty again. Therefore, we leave this to future work.

\paragraph{Novel reviews and meta-novelty.} We excluded reviews, along with comments, editorials and letters, since they do not usually propose a problem and an approach of their own. This is not to say that reviews cannot be novel. A useful review may reframe how a problem is understood, connect lines of work that were treated separately, or make explicit the assumptions that a field has taken for granted. Such contributions are novel at a different level: rather than offering a new problem or approach, they change how existing problems and approaches are seen. Our framework cannot capture this. Measuring this kind of novelty would require comparing how a field frames its problems before and after a review, which we leave to future work.

\paragraph{Summaries and terms.} The problem and approach statements are written by an LLM which highly affects the accuracy as the presence and absence of key terms can affect the distances a lot. The prompts ask for a detailed statement of each facet, which fixes the level of detail at which problems and approaches are described. What matters is that this level is consistent across papers: if the LLM describes some problems broadly and others narrowly, a broadly stated problem lies close to more papers than a narrowly stated one, and their scores are not comparable. Our experiments (Appendix ~\ref{app:nobel}) with abstracts in place of full texts illustrate this, as statements written from abstracts ).The choice of terms can also shift a statement's position: the same problem described with a specialised technical term or with a general paraphrase affect distances heavily. (Appendix ~\ref{app:blind_ext}).
 
\paragraph{Hindsight in the encoder.} The encoder learns which problems are similar from co-citations made after the papers it is applied to. When evaluating past years, it may therefore treat as similar two areas that were only connected later, possibly by the paper being evaluated, which could cause novel papers that bridge fields to be scored as not novel. This does not arise in prospective use, where the encoder is trained only on papers published before those it is applied to.

\paragraph{Soundness and false positives.} Our measure consider neither soundness nor utility. In the extreme, a paper that pursues a meaningless problem, or applies an approach that cannot work, may lie far from all earlier work and be scored as highly novel. A related kind of false positive concerns usefulness. A region may be sparse not because it is unexplored, but because few consider its problems worth studying. A paper there would be scored as novel, even though its sparsity reflects a lack of interest rather than a new direction. Such niche work is sometimes valuable, but sparsity alone does not show that it is. This also leaves the measure open to gaming: a researcher could raise their scores by choosing obscure problems or unusual combinations, without contributing anything of value.

\paragraph{Local density.} Papers in crowded areas of the literature are less likely to be flagged than papers in sparse areas, since their nearest neighbours lie closer. This is intended rather than a bias: our method treats sparsity as a proxy for novelty, and a paper in a crowded area follows a direction that many others are already taking. Density does, however, depend on how much of an area the corpus covers. An area may appear sparse because it is poorly represented in the PMC open-access subset rather than because few have worked on it, and papers there may be scored as more novel than they are.

\subsection{Intended Use}
We believe that novelty is one quality that good research may have, but not a measure of its value. Work that persists with an established approach is often what reveals its limits: anomalies become apparent only against the expectations that such work sets up \citep{kuhn1962structure}, and these limits in turn may prompt the search for alternatives. PCR is an example of this; cloning remained the method of choice and was productive for years before PCR was developed \citep{epo2003pcr}. Incremental work in a single direction can also accumulate into a substantial advance. 

One way we see this being useful is to gain perspective on how to tackle a problem. Because approaches are compared only among papers with similar problems, the neighbour set $\mathcal{K}_i$ of a paper lists the approaches previously taken to its problem, and papers with high approach novelty point to alternatives that the field has rarely pursued. Used in this way, the scores support researchers in surveying the options for a problem, rather than ranking them by how novel their work is.

\subsection{Code Availability}
The code is currently being prepared and will be released and provided here when ready.

\bibliography{refs}   
\pagebreak
\section{Appendix}
\subsection{Study on Nobel Prize}\label{app:nobel}
As an extension of the first experiment, we look at the work of Nobel laureates. We take six laureates whose prize-winning research
falls in biomedicine: Thomas R.\ Cech (Chemistry, 1989), Edwin G.\ Krebs (Physiology or Medicine, 1992), Eric
Wieschaus (1995), Stanley B.\ Prusiner (1997), G\"unter Blobel (1999) and Eric R.\ Kandel (2000). We chose them because
their publishing careers overlap the period of the HHMI corpus (1970--1995), so that corpus can be reused as the
comparison group. For Krebs, the prize-winning work (1955--1959) predates this period, so only his later papers are
included.

\paragraph{Laureate papers.} For each laureate, we collect from PubMed all journal articles published between 1970 and
1995 that list the laureate as an author (in any position) and have an abstract, excluding reviews, comments,
editorials and letters. This gives 865 papers (Cech 132, Krebs 154, Wieschaus 45, Prusiner 175, Blobel 204, Kandel
155), 761 of which were published before the laureate's prize year.

\paragraph{Abstract-based measures.} Only 69 of the 865 laureate papers are in the PMC open-access subset, so this
experiment uses titles and abstracts throughout: for the laureate papers, for their references, and for the corpus of
the HHMI experiment, whose problem and approach sentences we re-extract from the abstract instead of the full text
(abstracts from PubMed for 32{,}355 corpus papers and from the PMC record for 1{,}534; 5{,}029 have no abstract).
Extraction prompts, model and embedding are otherwise as in Section~\ref{method:inst}.

\paragraph{Candidate neighbours.} For a corpus paper, the candidate neighbours are the corpus papers published strictly
before it, excluding papers that share its last or corresponding author, as in the HHMI experiment. For a laureate
paper, the candidate neighbours are the corpus papers published strictly before it together with the paper's own
references (from OpenAlex), keeping only references that have a PubMed abstract, were published strictly before the
paper, are not reviews, comments, editorials or letters, and do not list the laureate as an author. Candidates
authored by the laureate or by the paper's last author are removed. $N_P$ and $N_A$ are then computed over the $k = 20$
nearest candidates by problem distance. For 745 of the 865 laureate papers, at least one of the 20 nearest neighbours is one of the paper's references (6.2 on average).

\paragraph{Thresholds and test.} Thresholds are set within each year as in the HHMI experiment, with $\tau_P(Y,q)$ and
$\tau_A(Y,q)$ the $(100-q)$th percentiles of $N_P$ and $N_A$ over the corpus papers of year $Y$ not authored by any of
the six laureates (33{,}757 papers between 1970 and 1995). Problem and approach novelty are defined as before, and we
report $q \in \{50, 20\}$. We compare the share of laureate papers flagged as novel with the share of corpus papers
flagged, using a two-sided Fisher's exact test.

\paragraph{Results.} Table~\ref{tab:nobel} reports the pooled rates and Table~\ref{tab:nobel-laureate} the rates for
each laureate. At $q = 50$, 73.2\% of laureate papers exhibit problem novelty, compared with 50.0\% of corpus papers
($p = 9.5 \times 10^{-43}$); at $q = 20$, the rates are 38.6\% and 20.0\% ($p = 3.6 \times 10^{-35}$). The problem
novelty rate exceeds the corpus rate for all six laureates at $q = 50$ (58.8\%--93.7\%) and for five of six at
$q = 20$ (Blobel: 17.6\%). For approach novelty, the rates are 11.0\% and 12.6\% at $q = 50$ ($p = 0.16$) and 10.1\%
and 9.1\% at $q = 20$ ($p = 0.34$). At $q = 20$, approach novelty is higher than in the corpus for Wieschaus (20.0\%,
$p = 0.019$) and Blobel (16.2\%, $p = 0.0013$). Counting either type, 84.2\% of laureate papers are novel vs.\ 62.6\%
of corpus papers at $q = 50$ ($p = 3.1 \times 10^{-43}$), and 48.7\% vs.\ 29.1\% at $q = 20$
($p = 1.0 \times 10^{-32}$). Without the papers' references among the candidate neighbours, the problem novelty rate of
laureate papers rises to 88.3\% at $q = 50$ and 64.4\% at $q = 20$, and the approach novelty rate falls to 5.8\% and
8.4\%.

\paragraph{Nobel Prize papers.} Interestingly, papers associated with Nobel Prize seem to exhibit problem novelty in contrast to HHMI papers. However, because the facets were extracted from abstracts, these results are not directly comparable with our other experiments, and we do not interpret them further.

\begin{table}[H]
\centering
\caption{Share of papers flagged as novel: laureate papers vs.\ corpus papers, 1970--1995. $p$: two-sided Fisher's
exact test.}
\label{tab:nobel}
\begin{tabular}{rlrrr}
\toprule
$q$ & Novelty & Laureates ($n = 865$) \% & Corpus ($n = 33{,}757$) \% & $p$ \\
\midrule
50 & Problem  & 73.2 & 50.0 & $9.5 \times 10^{-43}$ \\
   & Approach & 11.0 & 12.6 & 0.16 \\
   & Either   & 84.2 & 62.6 & $3.1 \times 10^{-43}$ \\
\midrule
20 & Problem  & 38.6 & 20.0 & $3.6 \times 10^{-35}$ \\
   & Approach & 10.1 &  9.1 & 0.34 \\
   & Either   & 48.7 & 29.1 & $1.0 \times 10^{-32}$ \\
\bottomrule
\end{tabular}
\end{table}

\begin{table}[H]
\centering
\caption{Share of each laureate's papers flagged as novel (\%). Corpus rates: problem 50.0 / 20.0, approach 12.6 / 9.1,
either 62.6 / 29.1 at $q = 50$ / $q = 20$.}
\label{tab:nobel-laureate}
\begin{tabular}{lrrrrrrrr}
\toprule
 & & & \multicolumn{3}{c}{$q = 50$} & \multicolumn{3}{c}{$q = 20$} \\
\cmidrule(lr){4-6}\cmidrule(lr){7-9}
Laureate & Prize & $n$ & Problem & Approach & Either & Problem & Approach & Either \\
\midrule
Cech      & 1989 & 132 & 90.9 &  3.8 & 94.7 & 49.2 &  7.6 & 56.8 \\
Krebs     & 1992 & 154 & 64.9 &  9.7 & 74.7 & 29.2 &  7.8 & 37.0 \\
Wieschaus & 1995 &  45 & 71.1 & 11.1 & 82.2 & 26.7 & 20.0 & 46.7 \\
Prusiner  & 1997 & 175 & 93.7 &  2.9 & 96.6 & 75.4 &  5.7 & 81.1 \\
Blobel    & 1999 & 204 & 58.8 & 19.6 & 78.4 & 17.6 & 16.2 & 33.8 \\
Kandel    & 2000 & 155 & 62.6 & 16.1 & 78.7 & 28.4 &  8.4 & 36.8 \\
\midrule
All       &      & 865 & 73.2 & 11.0 & 84.2 & 38.6 & 10.1 & 48.7 \\
\bottomrule
\end{tabular}
\end{table}

\subsection{Blind Test Extended Discussion}\label{app:blind_ext}

\paragraph{The cholera toxin case.} One of the 12 blind-test papers illustrates how sensitive the problem distance is
to the wording of the problem description. The paper (PMID 32176708) constructs a chimera of the
enterotoxigenic \emph{E.\ coli} (ETEC) adhesin CfaE with the B subunit of cholera toxin, which serves as an adjuvant, to
make the adhesin immunogenic. Its extracted descriptions are listed in Table~\ref{tab:cholera-text}.

The paper scores high whether its descriptions are extracted from the full text or from the abstract. The annotator, who reviewed the full-text version,
labelled its problem not novel and its approach novel, because none of the listed neighbours use cholera toxin. Neither
problem description mentions cholera toxin, which appears only in the approach descriptions. Among the paper's 20
nearest earlier neighbours (full text), one mentions cholera and three mention ETEC; most concern other bacterial
vaccines (LPS/O-antigen conjugates, outer membrane vesicles, porins), although 48 papers in the pool of 6{,}630 mention
cholera in their title or problem description.

We re-encoded the full-text problem description with ``such as the cholera toxin B subunit'' added after ``in the
absence of an adjuvant'', keeping the pool, filters and $k$ unchanged (Table~\ref{tab:cholera}). The mean problem
distance fell from 9.83 to 8.69, moving the paper from the 97th to the 73rd percentile of the cohort, and the number of
cholera papers among its 20 nearest neighbours rose from 1 to 10. The paper stays above the cohort median, but falls
out of the top 20\%. Because cholera toxin B is the paper's adjuvant, and thus part of its approach, the edit moves
approach content into the problem description. The test therefore shows how far a single phrase can move the problem
distance; it does not show which description is correct.

\begin{table}[H]
\centering
\small
\caption{The cholera toxin case (PMID 32176708): problem and approach descriptions extracted from the abstract and
from the full text, and the edited problem description used in Table~\ref{tab:cholera}.}
\label{tab:cholera-text}
\begin{tabular}{p{0.13\linewidth}p{0.80\linewidth}}
\toprule
\multicolumn{2}{l}{\textbf{Abstract}} \\
Problem  & The ETEC CFA/I adhesin CfaE is a promising vaccine antigen, but as a soluble antigen it is poorly
           immunogenic in the absence of an adjuvant, limiting its ability to elicit robust functional serum immune
           responses. \\
Approach & The researchers engineered a holotoxin-like dscCfaE-CTA2/CTB chimera by replacing the cholera toxin A1
           domain coding sequence with donor-strand-complemented CfaE in the cholera toxin operon, characterized its
           GM1-binding and hemagglutination activities, and compared its immunogenicity and functional serum responses
           in mice after intranasal or orogastric immunization against dscCfaE alone or admixed with CTB. \\
\midrule
\multicolumn{2}{l}{\textbf{Full text}} \\
Problem  & The soluble ETEC CFA/I adhesin CfaE is poorly immunogenic in the absence of an adjuvant, limiting its
           development as a component of a broadly protective multivalent ETEC vaccine. \\
Approach & The research approach was to construct a cholera toxin-like dscCfaE-CTA2/CTB chimera by replacing the CTA1
           domain with donor-strand-complemented CfaE, purify and biochemically characterize the assembled complex for
           GM1 binding, hemagglutination activity, and toxicity, and then immunize BALB/c mice intranasally and
           orogastrically to compare serum anti-CfaE/anti-CTB antibody and functional hemagglutination-inhibition
           responses elicited by the chimera versus dscCfaE alone, CTB alone, or a dscCfaE plus CTB admixture. \\
\midrule
\multicolumn{2}{l}{\textbf{Edited (full-text problem)}} \\
Problem  & The soluble ETEC CFA/I adhesin CfaE is poorly immunogenic in the absence of an adjuvant \emph{such as the
           cholera toxin B subunit}, limiting its development as a component of a broadly protective multivalent ETEC
           vaccine. \\
\bottomrule
\end{tabular}
\end{table}

\begin{table}[H]
\centering
\caption{The cholera toxin case (PMID 32176708): effect of naming the adjuvant in the full-text problem description.
Same pool (6{,}630 papers), filters and $k = 20$; percentile among the 100 cohort papers.}
\label{tab:cholera}
\begin{tabular}{lrrr}
\toprule
Problem description & $N_P$ & Percentile & Cholera papers among 20 nearest \\
\midrule
Original (extracted)                         & 9.83 & 97 &  1 \\
With ``such as the cholera toxin B subunit'' & 8.69 & 73 & 10 \\
\bottomrule
\end{tabular}
\end{table}

\subsection{Prompts}\label{app:prompts}

\paragraph{Problem.}
\begin{quote}\small\ttfamily
Read the paper in detail. Write out the research problem being addressed in one sentence. Ensure it is comprehensive yet concise and that how the problem is being addressed is not included.\\
<paper>\\
\{paper\}\\
</paper>
\end{quote}

\paragraph{Approach.}
\begin{quote}\small\ttfamily
Read the paper in detail. Write out the approach in one sentence. It should be written in a comprehensive yet concise manner. The sentence should solely describe the research approach.\\
<paper>\\
\{paper\}\\
</paper>
\end{quote}

\end{document}

%% file: macros.tex
\theoremstyle{plain}

\theoremstyle{definition}

\theoremstyle{remark}

\newcommand{\seclabel}[1]{\label{sec:#1}}

\definecolor{LightRed}{rgb}{1,0.9,0.9}
\definecolor{LightGreen}{rgb}{0.98,1.0,0.98}
\definecolor{DarkGreen}{rgb}{0.5,0.7,0.5}

\newcounter{questionCounter}

